\documentclass{article}

\PassOptionsToPackage{numbers}{natbib}
\usepackage[preprint]{neurips_2023}

\usepackage[utf8]{inputenc} 
\usepackage[T1]{fontenc}    
\usepackage{hyperref}       
\usepackage{url}            
\usepackage{booktabs}       
\usepackage{amsfonts}       
\usepackage{nicefrac}       
\usepackage{microtype}      
\usepackage{xcolor}         
\usepackage{subfig}
\usepackage{graphicx}
\usepackage{amsmath}
\usepackage{xspace}

\makeatletter
\DeclareRobustCommand\onedot{\futurelet\@let@token\@onedot}
\def\@onedot{\ifx\@let@token.\else.\null\fi\xspace}

\def\ie{\emph{i.e}\onedot}

\makeatother

\title{Explaining Explainability: Towards Deeper Actionable Insights into Deep Learning through Second-order Explainability}

\author{
E. Zhixuan Zeng \textsuperscript{1}\thefootnote{*}, Hayden Gunraj\textsuperscript{1}\thefootnote{*}, Sheldon Fernandez\textsuperscript{2}, Alexander Wong\textsuperscript{1}\\
 \textsuperscript{1} University of Waterloo, Canada\qquad
 \textsuperscript{2} DarwinAI Corp.\\
{\tt\small \{ezzeng, hayden.gunraj, alexander.wong\}@uwaterloo.ca}, \quad
{\tt\small sheldon@darwinai.ca}
}

\begin{document}

\maketitle

\def\thefootnote{*}\footnotetext{These authors contributed equally to this work.}

\begin{abstract}
  Explainability plays a crucial role in providing a more comprehensive understanding of deep learning models' behaviour. This allows for thorough validation of the model's performance, ensuring that its decisions are based on relevant visual indicators and not biased toward irrelevant patterns existing in training data. However, existing methods provide only instance-level explainability, which requires manual analysis of each sample. Such manual review is time-consuming and prone to human biases. To address this issue, the concept of second-order explainable AI (SOXAI) was recently proposed to extend explainable AI (XAI) from the instance level to the dataset level. SOXAI automates the analysis of the connections between quantitative explanations and dataset biases by identifying prevalent concepts.  In this work, we explore the use of this higher-level interpretation of a deep neural network's behaviour to allows us to "explain the explainability" for actionable insights. Specifically, we demonstrate for the first time, via example classification and segmentation cases, that eliminating irrelevant concepts from the training set based on actionable insights from SOXAI can enhance a model's performance.
\end{abstract}

\section{Introduction}
\label{sec:intro}

    \begin{figure}
        \subfloat[\label{fig:chainsaw_tsne}]{
            \includegraphics[width=\linewidth,trim={1cm 2cm 2cm 1.5cm},clip]{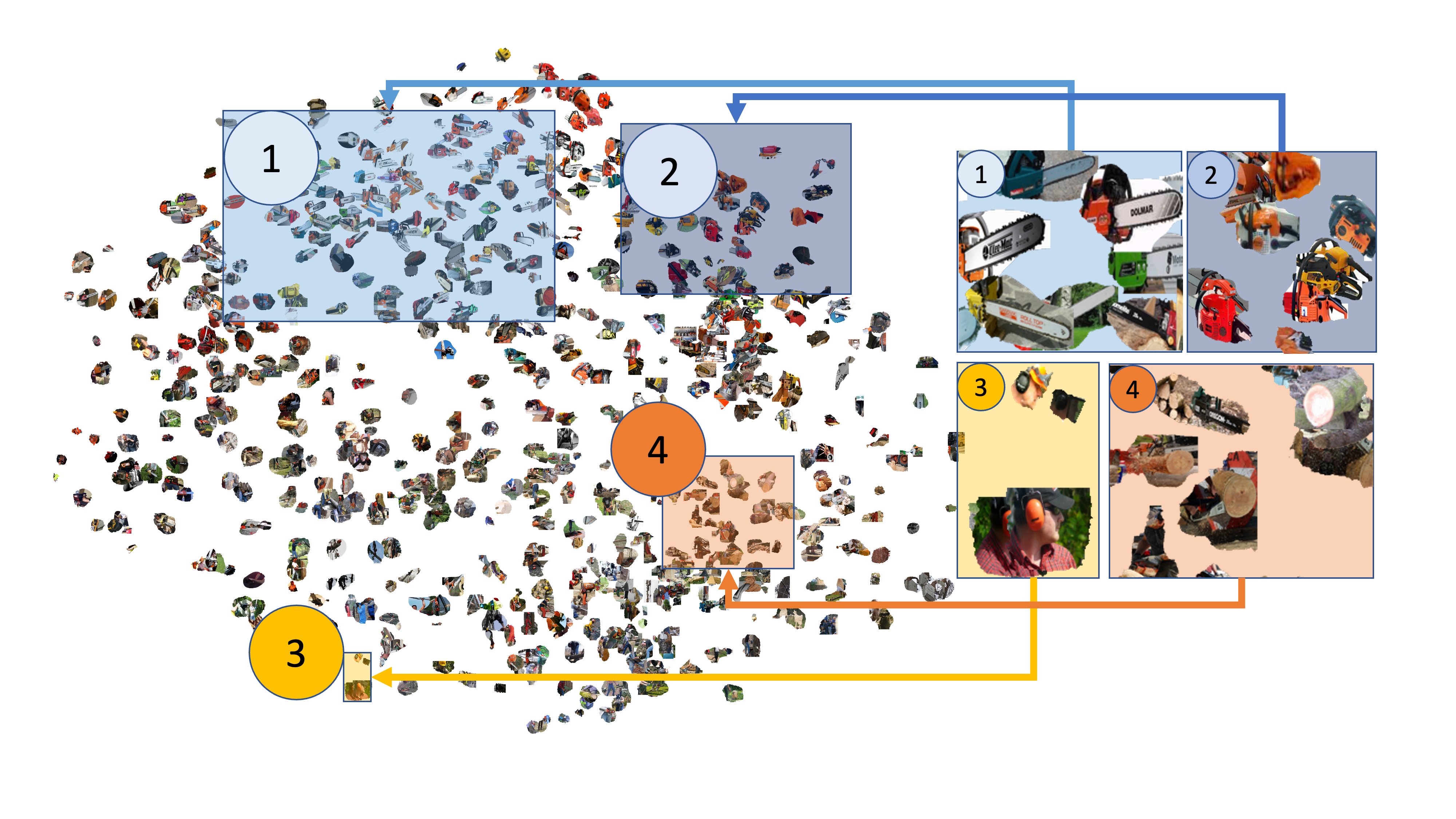}
        } \\
        \centering\subfloat[\label{fig:drills_tsne}]{
            \includegraphics[width=.95\linewidth,trim={2.5cm 2cm 5cm 3cm},clip]{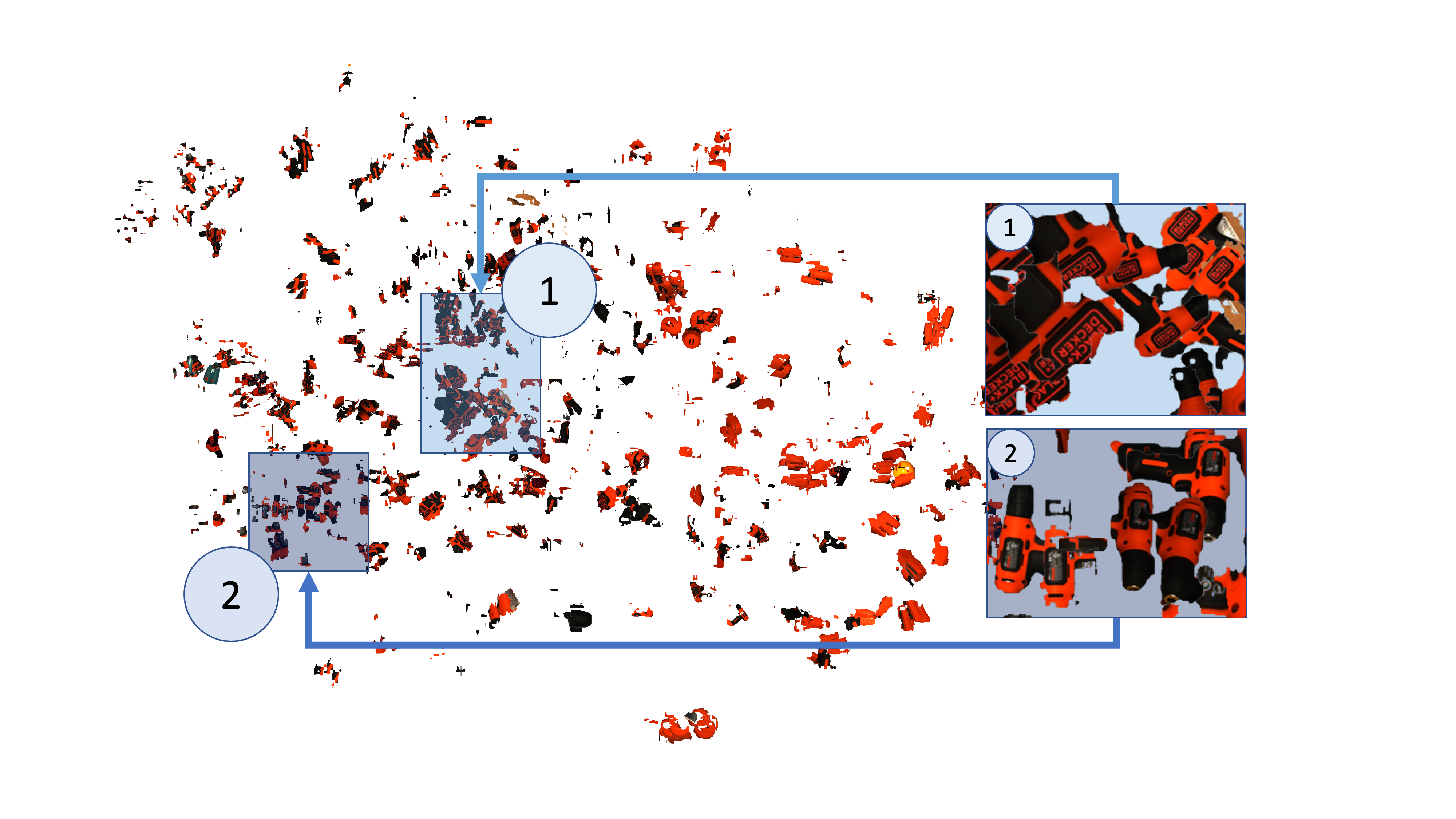}
        }
        \vspace{-8pt}
        \caption{SOXAI visualizations of a classification model on chainsaws \ref{fig:chainsaw_tsne} and a segmentation model on hand drills \ref{fig:drills_tsne}. Different regions show groupings of related quantitative explanations via first-order XAI, with significance discussed in Section \ref{sec:results}.}
        \label{fig:tsne_examples}
    \end{figure}

Although quantitative performance metrics such as accuracy are essential indicators of a deep neural network's performance, they do not offer insights into the decision-making process. To fill this gap in the performance analysis, explainable AI (XAI) can facilitate the auditing of model behaviour. This auditing helps ensure that the decisions are based on relevant visual indicators. Additionally, it can uncover potential biases in the training data, which may then be used to guide improvements to the training framework.

First-order explainability techniques such as Grad-CAM~\cite{gradcam}, integrated/expected gradients~\cite{intgrad,expgrad}, LIME~\cite{lime}, GSInquire~\cite{gsinquire}, and SHAP~\cite{shap} yield per-instance visualizations of explanations. However, reviewing these visualizations can be time-consuming, particularly for large-scale datasets with multiple classes or high intra-class variability. In addition, human biases can impact manual review. 

In this work, we explore the concept of second-order explainable AI (SOXAI)~\cite{soldernet} for obtaining actionable insights and demonstrate, for the first time, that such insights can be used to enhance model performance. SOXAI extends XAI from the instance level to the dataset level to enable the auditing of the model and dataset during development. Rather than relying on manual reviews of visual explanations to explore patterns in a model's decision-making behavior, SOXAI seeks to automatically unveil these patterns through the analysis of the relationships between quantitative explanations. This expedites the identification of the shared visual concepts utilized by a model during inference and can uncover apparent model and dataset biases. Furthermore, this improves transparency by uncovering problematic patterns  that exist among a groupings of examples in the dataset, which can adversely impact the model's decision-making process. In essence, SOXAI enables us to "explain the explainability" by providing higher-level interpretations of model behaviour for actionable insights.

\section{Methods}

\label{sec:methods}

The concept of SOXAI takes first-order instance-level quantitative explanations of samples in a dataset and groups similar embeddings of these explanations to generate a user-friendly visualization that enables the uncovering of patterns among different groupings of data to unveil trends.


Here, we employ GSInquire~\cite{gsinquire} to generate first-order quantitative explanations of a neural network's decision-making process across a dataset. GSInquire examines the network's activation signals in response to the input image and employs them to identify critical features within the sample that quantitatively led to the network's decision.  


\subsection{Second-order explainability}
Second-order explainability is treated as an embedding problem: given an image $I$ and the corresponding quantitative explanation $\alpha$ for the trained model $M$, we define the $n$\textsuperscript{th} element of the embedding $f:(I, \alpha)\rightarrow \mathbb{R}^N$ as:
\begin{equation} \label{main_theory}
    f(I,\alpha)_n = \frac{\sum_{i=1}^H\sum_{j=1}^W M(I)_{ijn}\alpha_{ij}}{\sum_{i=1}^H\sum_{j=1}^W\alpha_{ij}},
\end{equation}
\noindent producing an $N$-dimensional vector embedding from the regions of $I$ weighted by $\alpha$. Notably, $M$ is truncated such that its output is a convolutional feature map of size $H\times W\times N$, and $\alpha$ is resized to $H\times W$ to match. Equation \ref{main_theory} ignores regions not identified as critical and only considers regions with higher weighting score provided by $\alpha$ -- in essence, $f$ performs a weighted average of $M(I)$ with weights $\alpha$.

Here, we use t-distributed stochastic neighbour embedding (t-SNE)~\cite{tsne} to group the resulting embeddings across a full dataset~\cite{soldernet}. In addition, embeddings were reduced to 50 dimensions via principal component analysis before applying t-SNE to map them to a 2D space for visualization.



\section{Experimental Results and Discussion}
\label{sec:results}


We present two example cases of SOXAI visualization: image classification and foreground instance segmentation, discuss the actionable insights gained from each, and demonstrate how such actionable insights can be used to enhance model performance.

\textbf{Chainsaw classification:} To explore SOXAI for classification, we apply it to a ResNet-50 trained on ImageNet 1k~\cite{imagenet}. An example result for the chainsaw class can be seen in Figure \ref{fig:chainsaw_tsne}, which also highlights four groupings of interest. Groupings 1 and 2 show the frontal part of chainsaws (\ie, the cutting chain and guide bar) and the handle, respectively, demonstrating that the model has learned important features representing the target class. However, smaller groupings highlighted in areas 3 and 4 also reveal biases that the model has learned over time. In grouping 3, we see that the model has learned a relationship between earmuffs commonly worn when using chainsaws and the actual class prediction. Grouping 4 shows images of logs and even wooden sculptures instead of chainsaws directly.

Through the use of SOXAI, we were able to quickly identify reoccurring biases learned by the model towards objects that commonly appear in the same frame as the target class. This was accomplished without the need to manually inspect each example in the validation set, as would be necessary for first-order XAI algorithms. Based on the identified biases, enhanced model performance may be achieved by better-targeted elimination of biases in future training and data collection or cleaning.

\begin{figure}
    \centering
    \subfloat[\label{fig:drill_incomplete}]{
        \includegraphics[width=0.5\linewidth]{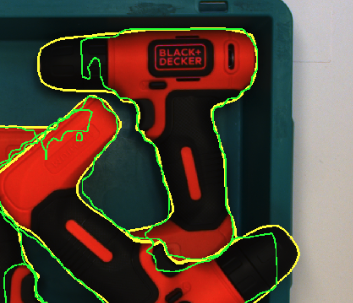}
    }
    \subfloat[\label{fig:drill_filled}]{
        \includegraphics[width=0.5\linewidth]{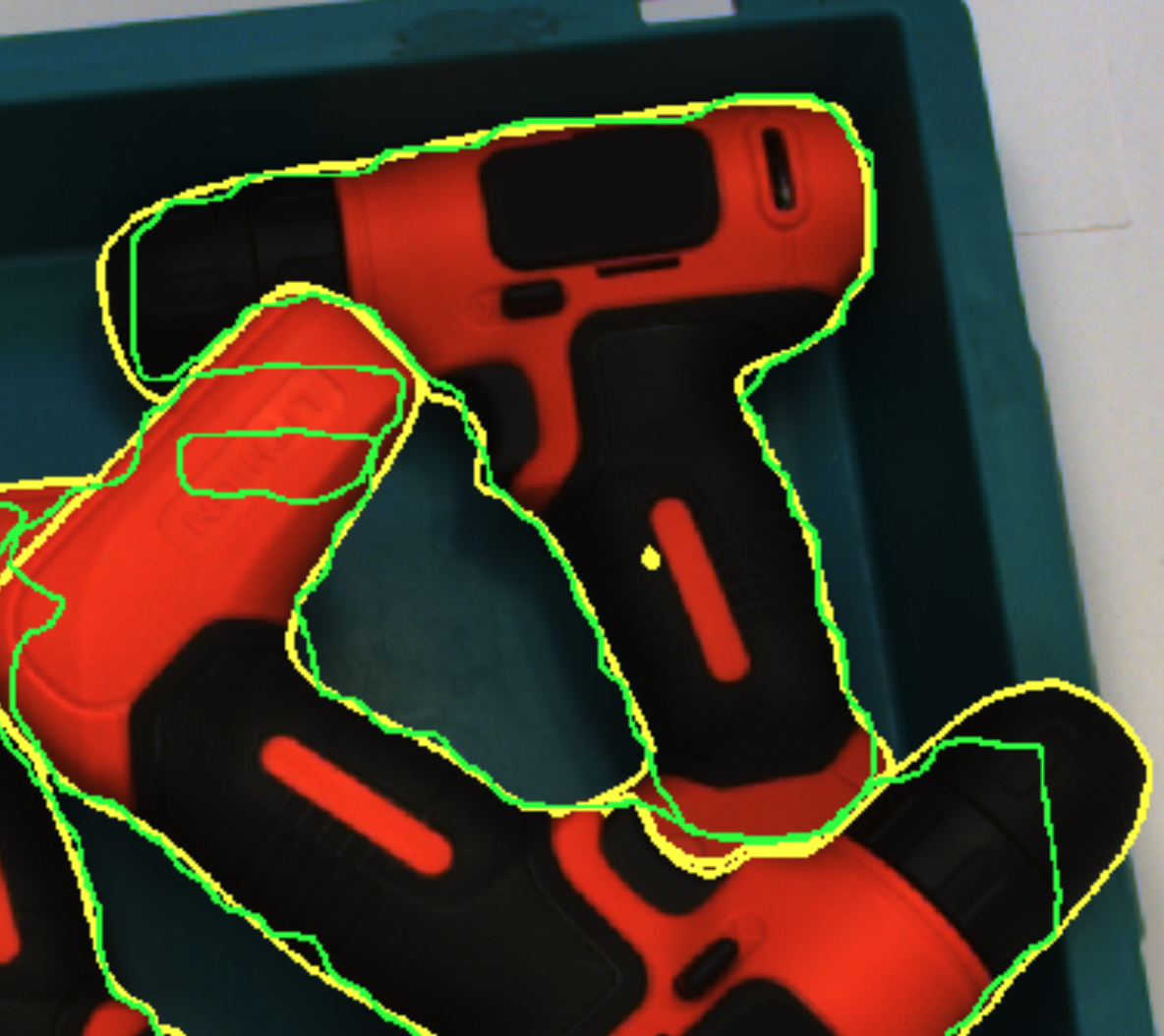}
    }
    
    \caption{Example of a drill with incomplete segmentation \ref{fig:drill_incomplete}, and with the label filled in \ref{fig:drill_filled}. Model prediction is outlined in green.}
    \label{fig:drill_example}
\end{figure}

\textbf{Drill segmentation:} Here, we apply SOXAI to a MaskRCNN model \cite{he2017maskrcnn} trained on the MetaGraspNet dataset \cite{chen2021metagraspnet} to detect foreground objects. As an example, we analyze the segmentation of drills, an object category not seen in the training set, chosen for its geometric and textural complexity. Figure \ref{fig:drills_tsne} presents the SOXAI result, highlighting two groupings representing different faces of the drill.

The face shown in grouping 1 exhibits a high level of focus on the large logo. Since the model was not explicitly trained to recognize drills, some other foreground object must have biased it towards recognizing letters. We observe that the large logo is over-represented in the grouping, while the frontal black head of the drill is underrepresented.

To investigate further, we evaluate the prevalence of incomplete segmentations of the drill when each face is visible, such as the incomplete segmentation shown in Figure \ref{fig:drill_incomplete}. We find that 37\% of predictions for drills with the large logo facing up are incomplete segmentations, with much of the frontal black segment missing, while only 14\% of segmentation predictions on the other face are incomplete.

To confirm the model's bias towards text, we mask out the logo (see Figure \ref{fig:drill_filled}), and evaluate the mAP score. We observe an increase from 0.592 to 0.618, suggesting that allowing the model to ignore its learned bias and focus on a fuller representation of the object improves its performance.  These example cases demonstrate the usefulness  of SOXAI for unveiling actionable insights into model biases that can be used to enhance a model's performance.



{\small
\bibliographystyle{ieee_fullname}
\bibliography{egbib}

\begin{thebibliography}{10}\itemsep=-1pt

\bibitem{chen2021metagraspnet}
Yuhao Chen, E.~Zhixuan Zeng, Maximilian Gilles, and Alexander Wong.
\newblock Metagraspnet: a large-scale benchmark dataset for vision-driven
  robotic grasping via physics-based metaverse synthesis.
\newblock {\em arXiv preprint arXiv:2112.14663}, 2021.

\bibitem{imagenet}
J. {Deng}, W. {Dong}, R. {Socher}, L. {Li}, {Kai Li}, and {Li Fei-Fei}.
\newblock Imagenet: A large-scale hierarchical image database.
\newblock In {\em 2009 IEEE Conference on Computer Vision and Pattern
  Recognition (CVPR)}, pages 248--255, 2009.

\bibitem{expgrad}
Gabriel Erion, Joseph~D. Janizek, Pascal Sturmfels, Scott~M. Lundberg, and
  Su-In Lee.
\newblock Improving performance of deep learning models with axiomatic
  attribution priors and expected gradients.
\newblock {\em Nature Machine Intelligence}, 3:620–631, 2021.

\bibitem{soldernet}
Hayden Gunraj, Paul Guerrier, Sheldon Fernandez, and Alexander Wong.
\newblock {SolderNet}: Towards trustworthy visual inspection of solder joints
  in electronics manufacturing using explainable artificial intelligence.
\newblock In {\em 35th Annual Conference on Innovative Applications of
  Artificial Intelligence ({IAAI}-23)}. Association for the Advancement of
  Artificial Intelligence ({AAAI}), 2023.

\bibitem{he2017maskrcnn}
Kaiming He, Georgia Gkioxari, Piotr Doll{\'a}r, and Ross Girshick.
\newblock Mask r-cnn.
\newblock In {\em Proceedings of the IEEE international conference on computer
  vision}, pages 2961--2969, 2017.

\bibitem{gsinquire}
Zhong~Qiu Lin, Mohammad~Javad Shafiee, Stanislav Bochkarev, Michael St.~Jules,
  Xiao~Yu Wang, and Alexander Wong.
\newblock Do explanations reflect decisions? a machine-centric strategy to
  quantify the performance of explainability algorithms.
\newblock {\em arXiv preprint. arXiv:1910.07387}, 2019.

\bibitem{shap}
Scott Lundberg and Su-In Lee.
\newblock A unified approach to interpreting model predictions.
\newblock In {\em 30th International Conference on Neural Information
  Processing Systems (NIPS 2017)}, page 768–4777, 2017.

\bibitem{lime}
Marco Ribeiro, Sameer Singh, and Carlos Guestrin.
\newblock {``}why should {I} trust you?{''}: {E}xplaining the predictions of
  any classifier.
\newblock In {\em Proceedings of the 2016 Conference of the North {A}merican
  Chapter of the Association for Computational Linguistics: Demonstrations},
  pages 97--101, San Diego, California, June 2016. Association for
  Computational Linguistics.

\bibitem{gradcam}
R.~R. {Selvaraju}, M. {Cogswell}, A. {Das}, R. {Vedantam}, D. {Parikh}, and D.
  {Batra}.
\newblock Grad-cam: Visual explanations from deep networks via gradient-based
  localization.
\newblock In {\em 2017 IEEE International Conference on Computer Vision
  (ICCV)}, pages 618--626, 2017.

\bibitem{intgrad}
Mukund Sundararajan, Ankur Taly, and Qiqi Yan.
\newblock Axiomatic attribution for deep networks.
\newblock In {\em Proceedings of the 34th International Conference on Machine
  Learning - Volume 70}, ICML'17, page 3319–3328. JMLR.org, 2017.

\bibitem{tsne}
Laurens van~der Maaten and Geoffrey Hinton.
\newblock Visualizing data using t-sne.
\newblock {\em Journal of Machine Learning Research}, 9(86):2579--2605, 2008.

\end{thebibliography}
}

\end{document}